\documentclass[letterpaper, 10pt, conference]{ieeeconf}      

\IEEEoverridecommandlockouts                              

\usepackage[dvipsnames, table]{xcolor}
\usepackage{pgf}
\usepackage{pgfplots}
\pgfplotsset{compat=1.14}
\usepackage{float}
\usepackage{booktabs}
\usepackage{siunitx}
\DeclareSIUnit{\nothing}{\relax}
\usepackage{multirow}
\usepackage{nicefrac}
\usepackage[hidelinks]{hyperref}

\definecolor{somegray}{rgb}{0.5, 0.5, 0.5}
\newcommand{\darkgrayed}[1]{\textcolor{somegray}{#1}}
\makeatletter
\newcommand*\titleheader[1]{\gdef\@titleheader{#1}}
\AtBeginDocument{%
  \let\st@red@title\@title
  \def\@title{%
    \vskip-2.0em
    \bgroup\normalfont\large\centering\@titleheader\par\egroup
    \vskip0.0em\st@red@title}
}

\makeatother

\titleheader{\darkgrayed{This paper has been accepted for publication in the IEEE ICRA 2023 conference\\\copyright 2023 IEEE.}}

\DeclareSIUnit{\nothing}{\relax}
\DeclareSIUnit\mac{MAC}

\title{\LARGE \bf
Deep Neural Network Architecture Search for Accurate Visual Pose Estimation aboard Nano-UAVs
}

\author{E. Cereda$^{1}$, L. Crupi$^{2}$, M. Risso$^{2}$, A. Burrello$^{3}$, L. Benini$^{34}$, A. Giusti$^{1}$, D. Jahier Pagliari$^{2}$, and D. Palossi$^{14}$
\thanks{This work was partially supported by the Secure Systems Research Center (SSRC) of the UAE Technology Innovation Institute (TII) and the Swiss National Science Foundation (SNSF) through the NCCR Robotics.}
\thanks{$^{1}$E. Cereda, A. Giusti, and D. Palossi are with the Dalle Molle Institute for Artificial Intelligence, USI and SUPSI, Lugano, 6962, Switzerland
        {\tt\small name.surname@idsia.ch}}%
\thanks{$^{2}$L. Crupi, M. Risso, and D. Jahier Pagliari are with the
Department of Control and Computer Engineering, Politecnico di Torino, Turin, 10129, Italy
        {\tt\small name.surname@polito.it}}%
\thanks{$^{3}$A. Burrello and L. Benini are with the Department of Electrical, Electronic, and Information Engineering, University of Bologna, Bologna, 40136, Italy
        {\tt\small name.surname@unibo.it}}%
\thanks{$^{4}$L. Benini and D. Palossi are also with the Integrated Systems Laboratory, ETH Z\"urich, Z\"urich, 8092, Switzerland
        }%
}

\begin{document}

\maketitle

\begin{abstract}
Miniaturized autonomous unmanned aerial vehicles (UAVs) are an emerging and trending topic. 
With their form factor as big as the palm of one hand, they can reach spots otherwise inaccessible to bigger robots and safely operate in human surroundings.
The simple electronics aboard such robots (sub-\SI{100}{\milli\watt}) make them particularly cheap and attractive but pose significant challenges in enabling onboard sophisticated intelligence.
In this work, we leverage a novel neural architecture search (NAS) technique to automatically identify several Pareto-optimal convolutional neural networks (CNNs) for a visual pose estimation task.
Our work demonstrates how real-life and field-tested robotics applications can concretely leverage NAS technologies to automatically and efficiently optimize CNNs for the specific hardware constraints of small UAVs.
We deploy several NAS-optimized CNNs and run them in closed-loop aboard a 27-g Crazyflie nano-UAV equipped with a parallel ultra-low power System-on-Chip.
Our results improve the State-of-the-Art by reducing the in-field control error of 32\% while achieving a real-time onboard inference-rate of $\sim$\SI{10}{\hertz}@\SI{10}{\milli\watt} and $\sim$\SI{50}{\hertz}@\SI{90}{\milli\watt}.
\end{abstract}

\section*{Supplementary video material}
In-field tests: \url{https://youtu.be/dVCScckvcg8}.

\section{Introduction} \label{sec:intro}

Nano-sized unmanned aerial vehicles (UAVs) are gaining significant momentum due to their reduced form factor (sub-\SI{10}{\centi\meter}) and weight (sub-\SI{40}{\gram}), which allows them to fulfill sensitive missions, such as in GPS-denied narrow spaces and human proximity.
Moreover, the simplified sensory, mechanical, and computational sub-systems available aboard these platforms make them particularly cheap and attractive compared to their bigger counterparts.
However, making them fully autonomous, i.e., no external/remote infrastructure or computation, is still challenged by their simplified electronics, e.g., sub-\SI{100}{\milli\watt} compute power, which is 1-2 orders of magnitude less than mobile phones' processors.

These constraints hinder the adoption of standard algorithmic pipelines, which often rely on memory/compute-intensive sophisticated planning/localization/mapping algorithms, and heavy pre-trained perception convolutional neural networks (CNNs).
Therefore, running these methods aboard nano-UAVs is unfeasible.
In this context, tiny CNNs increasingly define the State-of-the-Art (SoA) of autonomous nano-drones~\cite{ai-deck,frontnet}.
First, they can be trained to operate on data from very limited sensors, such as ultra-low power, and tiny cameras with poor resolution/dynamic range.
Second, CNNs have predictable and fixed computational and memory requirements at inference time. 
Third, the system designer can tune such requirements to fit the available resources on the robot by defining a suitable network architecture.

Roboticists integrating CNNs in larger robots can often overlook the third aspect, relying on standard SoA architectures such as ResNet50~\cite{resnet50} that easily fit in the available resources.
In contrast, thoroughly defining a suitable custom CNN is crucial when developing nano-class robot systems.  
The choice of the neural network architecture determines whether the model can run on the robot and directly impacts two critical parameters that affect the robot's behavior: \textit{i}) prediction performance and \textit{ii}) real-time throughput.

For many years, the only approach to fulfill the requirements posed by this complex optimization scenario with contrasting objectives (i.e., obtaining accurate yet deployable CNNs) was to resort to a manual, tedious, error-prone, and time-consuming iterative hyper-parameters tuning based on heuristics and rules-of-thumb.
Nowadays, the \textit{go-to} approach to perform such optimization is based upon the neural architecture search (NAS) paradigm~\cite{nas_ea, mnasnet}.
NAS tools enable an automatic exploration over an arbitrarily large search space of different network topologies and hyper-parameters settings.
Furthermore, many novel NAS approaches can directly optimize complex cost functions by combining different objectives~\cite{mnasnet, proxylessnas, pitjournal}, such as regression performance and the network's computational complexity.

In this work, we exploit a novel computationally efficient NAS technique~\cite{pitjournal}, able to generate Pareto-optimal CNN architectures in the accuracy vs. model size, to optimize a vision-based human pose estimation task.
First, we contribute by enhancing the functionalities of an existing NAS engine, which is needed to explore two different seed CNNs: PULP-Frontnet~\cite{frontnet} and MobileNetv1~\cite{mobilenets}.
Then, we thoroughly analyze and deploy multiple CNNs on our target nano-drone robotic platform.
Ultimately, the delivered CNNs improve the SoA baseline~\cite{frontnet} at least for one metric among size (up to $5.6\times$ smaller), speed (up to $1.5\times$ faster), and accuracy (up to 32\% lower horizontal displacement error). 
The improvement in accuracy is finally confirmed by a challenging in-field testing setup, with a \textit{never-seen-before} environment.

\section{Related work} \label{sec:related_work}

\textbf{Neural architecture search.}
NAS tools assist designers in the design phase of DNNs by automatically exploring a large space of architectures defined as combinations of different layers and/or hyper-parameters.
On constrained platforms, these tools usually minimize an objective function that depends both on task accuracy and non-functional cost metrics (e.g., memory footprint, latency, or energy consumption). 
Early NAS algorithms were based on \textit{evolutionary algorithms} (EA)~\cite{nas_ea} and reinforcement learning (RL)~\cite{mnasnet}.
These methods can explore arbitrary search spaces, and optimize any cost function by iteratively sampling a network, training it to convergence to evaluate performance, and then using this information to drive the following sampling.

However, their extreme flexibility implies poor scalability, due to their computational requirement, i.e., thousands of GPU hours, even for simple tasks.
Instead, differentiable NAS (DNAS)~\cite{darts, proxylessnas} has been proposed to mitigate this issue.
Early DNASes exploit \textit{supernets}, i.e., networks whose layers' outputs are obtained as a weighted sum of multiple alternative sub-layers (e.g., different types of convolution) applied to the same input~\cite{darts}.
The weights assigned to each alternative are optimized \textit{during training}, based on an accuracy/complexity-dependent loss, and the final network is obtained by selecting, for each layer, the alternative associated with the largest weight at the end of the search.
Compared to EA/RL approaches, DNAS trades some flexibility in the definition of the optimization target, which must be differentiable, in exchange for a search process that requires a \textit{single training}.
However, the convergency of a supernet is \textit{i}) not trivial, and \textit{ii}) still expensive (in terms of GPU memory), since multiple alternatives are instantiated for each layer.

\textit{Mask-based DNAS} are a further step toward lightweight architecture search.
They replace the supernet with a single-path DNN, usually referred to as \textit{seed}~\cite{morphnet, wan2020fbnetv2, pitjournal}, and their search space is composed of sub-architectures obtained from the seed \textit{by subtraction} (e.g., eliminating some channels in each convolution).
In practice, these sub-networks are simulated at training time using \textit{trainable masks}, which prune part of the seed. 
As in supernet DNASes, the masks are optimized during training, but since the seed is much smaller than a supernet, the time and memory overhead compared to regular DNN training is minimal.
On the one hand, mask-based DNASes can only produce networks derived from the seed; on the other, this allows for a much more fine-grained search space exploration. 
For instance, they can easily generate convolutions with an arbitrary number of channels (e.g., 17, 25, or 31 for a 32-channel seed): prohibitive with a supernet approach~\cite{pitjournal}.

\textbf{Nano-drones.}
Deploying CNN models for autonomous navigation on a nano-drone introduces severe sensorial, computational, and memory constraints due to limited form factor, payload, and energy availability.
Solutions built around commercial off-the-shelf (COTS) microcontroller units (MCUs)~\cite{neuralswarm,uwbbias} can afford only minimal neural architectures to achieve real-time performance, such as~\cite{uwbbias} with \SI{27}{\kilo\mac} (multiply-accumulate operations) per frame at \SI{100}{\hertz}.
These approaches are thus suitable only for low-dimensional input signals, not for processing camera images.
More advanced visual approaches~\cite{frontnet,pulp-dronet} have been enabled by careful hardware-software co-design, exploiting general-purpose ultra-low-power multi-core System-on-Chip (SoC)~\cite{pulp}, integrated software deployment pipelines~\cite{conti2020nemo,burrello2020dory} and manually-tuned CNN architectures.
These technological breakthroughs enabled PULP-Frontnet~\cite{frontnet}, a model with \SI{14.7}{\mega\mac} (3 order of magnitude larger than~\cite{uwbbias}), to achieve an inference rate of \SI{48}{\hertz} while consuming \SI{96}{\milli\watt}, aboard a COTS Crazyflie 2.1 nano-drone.

An open research question still revolves around CNN's complexity/memory reduction, as answering it would pave the way toward \textit{i}) better energy utilization on battery-limited devices, and \textit{ii}) enabling multi-tasking systems, yet not reached on nano-drones.
Recent works~\cite{navardi22optim,tiny-dronet} have aimed at reducing CNNs' memory footprint with minimal degradation in accuracy, achieving up to $27\times$ reductions in computation with just 4\% lower accuracy~\cite{tiny-dronet}, but still at the cost of extensive manual fine-tuning of the architectures. 
In this work, we leverage NAS techniques to efficiently and automatically cope with this problem in a robotic domain.

\textbf{Human pose estimation.}
We consider the robotic problem of monocular relative pose estimation of a human subject from an autonomous nano-drone. 
Despite the remarkable accuracy of SoA computer vision approaches~\cite{Sun_2018_ECCV,luvizon18multitask,dense-pose}, they are still far from the reach of nano-drones due to the required amount of computational resources~\cite{dense-pose} ($\sim$10s \si{\giga\mac}).
To date, the PULP-Frontnet CNN~\cite{frontnet} represents one of the few examples of a human pose estimation task fully executed aboard a Crazyflie nano-drone in real-time.
While complex CNNs manage to estimate entire skeletons~\cite{luvizon18multitask} or even dense 3D mesh representations~\cite{dense-pose} of the person, PULP-Frontnet minimizes its prediction to a 3D point in space ($x, y, z$) and a rotation angle w.r.t. the gravity z-axis $(\phi)$. 
MobileNetv1~\cite{mobilenets} is another SoA CNN vastly used for vision-based classification tasks, which has also been proven accurate (e.g., 68.2\% top-1 accuracy on ImageNet) when streamlined down to the same power envelope allotted on our nano-drone~\cite{capotondi2020cmix,burrello2020dory}.
For these reasons, we choose PULP-Frontnet and MobileNetv1 as seed models for our NAS, and we will refer to the PULP-Frontnet model as the SoA baseline (F$_{\mathit{SoA}}$) for the in-field comparison.
\section{Methodology} \label{sec:methodology}

\subsection{Seed deep neural networks} \label{sec:DNNs}

In this work, we use a mask-based DNAS tool to optimize three seed models: one based on the shallow PULP-Frontnet architecture, and two based on a deeper MobileNetv1.
PULP-Frontnet is composed by 6 convolutional layers requiring \SI{304}{\kilo\nothing} parameters and up to \SI{14.7}{\mega\mac}.
The design of the network was first introduced in~\cite{frontnet} specifically for deployment on a nano-drone.
On the other hand, MobileNetv1~\cite{mobilenets} is a deeper network formed by 27 different convolutional layers, which mostly differ from the PULP-Frontnet by the separable depthwise and pointwise convolutional layers instead of traditional ones. 
Our two MobileNetv1 seed networks vary by width multiplier, a hyperparameter that controls the size of the feature maps throughout the model: we consider width multipliers of 1.0 (M$^{1.0}$) and 0.25 (M$^{0.25}$), corresponding to respectively \SI{3.21}{\mega\nothing} and \SI{204}{\kilo\nothing} parameters.

\subsection{Network architecture search} \label{sec:NAS}

\begin{figure}[t]
  \includegraphics[width=\columnwidth]{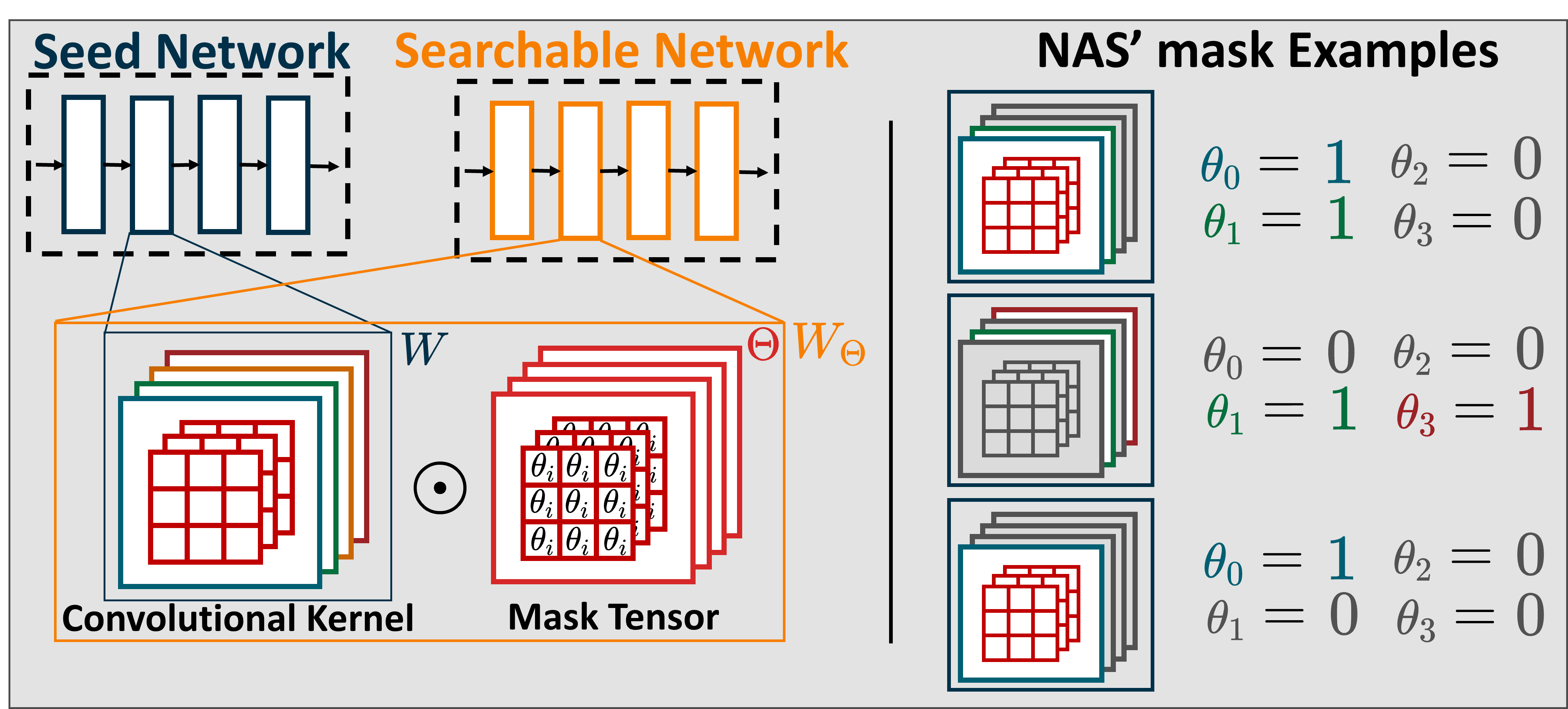}
  \caption{Left: the proposed masking scheme. Right: three possible outcomes of the channel search process on a four-filters layer. $\theta_{0}=\theta_{1}=1$ and $\theta_{2}=\theta_{3}=0$ means that the first two channels are kept alive, while the second two are removed. Only the second and last channels are kept alive in the second example, and only the first channel in the last one.}
  \label{fig:ch_search}
\end{figure}

We optimize the network architecture using a mask-based DNAS tool, namely \textit{pruning in time} (PIT)~\cite{pitjournal}, whose scope is to search for smaller and more lightweight networks which hold almost the same accuracy as the seed models.
PIT shows low memory and execution time overheads (only $\sim$30\% of the training time), several orders of magnitude lower than other NAS engines~\cite{nas_ea, mnasnet, darts}.
Additionally, the availability of good reference architectures for our task that can be used as \textit{seed} fits well with a mask-based approach, which efficiently explores different sub-networks contained within the seed.

The NAS we propose performs a fine-grained search of the number of output channels in all convolutional layers of a 2D CNN. 
The approach is derived from the channel search scheme originally described in~\cite{pitjournal} for 1D Temporal Convolutional Networks.
Figure~\ref{fig:ch_search} summarizes our NAS functionality: for each convolutional layer in the seed, its weight tensor $W$, with $C_{out}$ output channels, is isolated, and the correspondent searchable tensor $W_{\Theta}$ is built as: 

\begin{equation} \label{eq:search_w}
W_{\Theta} = W \odot \mathcal{H}(\theta)
\end{equation}

where $\odot$ is the Hadamard product, $\theta$ is a trainable mask vector of $C_{out}$ elements, in which each component $\theta_{i}$ represents the mask for a specific channel, and $\mathcal{H}$ is the Heaviside step function used to binarize $\theta$.
Depending on the binarized values of $\mathcal{H}(\theta_{i})$ the correspondent \textit{i}-th output channel of $W$ may be kept alive ($\mathcal{H}(\theta_{i})=1$) or removed ($\mathcal{H}(\theta_{i})=0$).
Therefore PIT finds models which only vary in the number of channels w.r.t. the seed networks.
Compared to its original version proposed in~\cite{pitjournal}, we also added support for \textit{jointly exploring} the channels of pointwise and depthwise layers, the two main layers in depth-separable convolutional blocks and basic components of MobileNet architectures.
To do so, we used shared masks to select the number of channels of these two layers, since the output channels of a depthwise convolution are completely determined by the preceding pointwise layer.

The obtained network is inserted in a normal training loop where $W$ and $\theta$ are trained in conjunction to solve the following optimization problem:

\begin{equation} \label{eq:opt_dnas}
\min_{W, \theta} \mathcal{L}(W; \theta) + \lambda \mathcal{R}(\theta)
\end{equation}

where $\mathcal{L}$ is the task-specific loss function (i.e., the same loss function as the seed CNNs) and $\mathcal{R}$ is a regularization loss term representing a cost metric to be optimized (e.g., n. of parameters, n. of operations, etc.).
$\lambda$ is the so-called \textit{regularization strength}, a hand-tuned scalar term used to balance the two losses.
Once loss terms of~Eq.\ref{eq:opt_dnas} are defined, $\lambda$ represents the main knob upon which the designer can act to drive the search towards more accurate or less costly networks.
For our use case, we considered $\lambda \in [5\cdot10^{-11}:5\cdot10^{-5}]$.
As regularization loss $\mathcal{R}$, we used a differentiable estimate of the number of parameters of the network as a function of the NAS mask values. 
In this way, we can bias the exploration phase towards architectures that are both small and accurate.

\subsection{System design} \label{sec:system_design}

\textbf{Robotic platform.} 
Our target robotic platform is the Bitcraze Crazyflie 2.1\footnote{https://www.bitcraze.io/products/crazyflie-2-1/}, a \SI{27}{\gram} nano-quadrotor.
This drone features a main STM32 MCU in charge of basic functionalities, such as state estimation and control loops.
In our in-field deployment, it is extended with a \SI{5}{\gram} commercial AI-deck companion board~\cite{ai-deck}.
The AI-deck features an additional MCU: the GreenWaves Technologies GAP8, which embodies the parallel ultra-low power paradigm~\cite{pulp} through a RISC-V-based multi-core SoC.
These two processors communicate via a bidirectional UART interface.
The GAP8 is designed with two power domains: a single-core \textit{fabric controller} (FC) that orchestrates the interaction with external memories/sensors and offloads computationally intensive kernels on a second 8-core \textit{cluster} (CL) domain. 
The SoC's memory hierarchy relies on \SI{64}{\kilo\byte} of low-latency L1 memory shared among all cluster cores and \SI{512}{\kilo\byte} of L2 memory within the FC domain. 
The GAP8 also features two DMA engines to efficiently automate data transfers from/to all memories and external peripherals, such as \SI{8}{\mega\byte} off-chip DRAM and a QVGA monochrome camera, both available on the AI-deck.
However, it provides neither data caches nor hardware floating-point units, dictating explicit data management and the adoption of integer-quantized arithmetic, respectively.

\textbf{Deployment tools.} 
To fulfill the platform's requirements, we exploit two tools to \textit{i}) train integer-only neural networks and \textit{ii}) automatically and optimally generate C code for the target network.
The first tool, the open-source NEMO library~\cite{conti2020nemo}, based on the PyTorch framework, is used to train the network in three sequential steps.
First, NEMO trains a \textit{full-precision} floating-point network to minimize the sum of the L1 loss for the pose estimation vector ($x,y,z,\phi$).
We use the SGD optimizer with a \textit{lr} of 0.001 over 100 epochs, selecting at the end the model which achieved the lowest validation loss. 
After, we convert this model into a \textit{fake-quantized} network.
In this stage, weights and activations are still represented as float32 values, but their magnitude can assume only a discrete (256 for 8-bits quantization) set of values.
Given the support offered by the optimized kernels for GAP8, we use linear per-layer quantization and PACT~\cite{choi2018pact} in this fine-tuning step.
We choose to use 8-bit quantization since it is natively supported by the SIMD operations in the RISC-V ISA extensions, which allow the execution of four int8 MACs within a single cycle.
To maximize the accuracy of the network, we initialize the fake-quantized network with the floating point weights, and we perform 100 additional train epochs with the same parameters.
The final step is the creation of the \textit{integer deployable} network.
Compared to the fake-quantized network, all the tensors are represented by integer numbers and a float scale factor.
Tensors $\mathbf{t}$ are approximated as 
\begin{equation}
    \mathbf{t} \approx \varepsilon_\mathbf{t}\cdot \mathbf{t^*} \label{eq:1}\;, 
\end{equation}
where $\mathbf{t}$ is the fake-quantized tensor, $\mathbf{t^*}$ is the integer-only tensor, and $\varepsilon_\mathbf{t}$ is the scale floating point factor.
Therefore, the network can run entirely in the integer domain by multiplying and accumulating only integer values.

\begin{figure}[t]
  \includegraphics[width=\columnwidth]{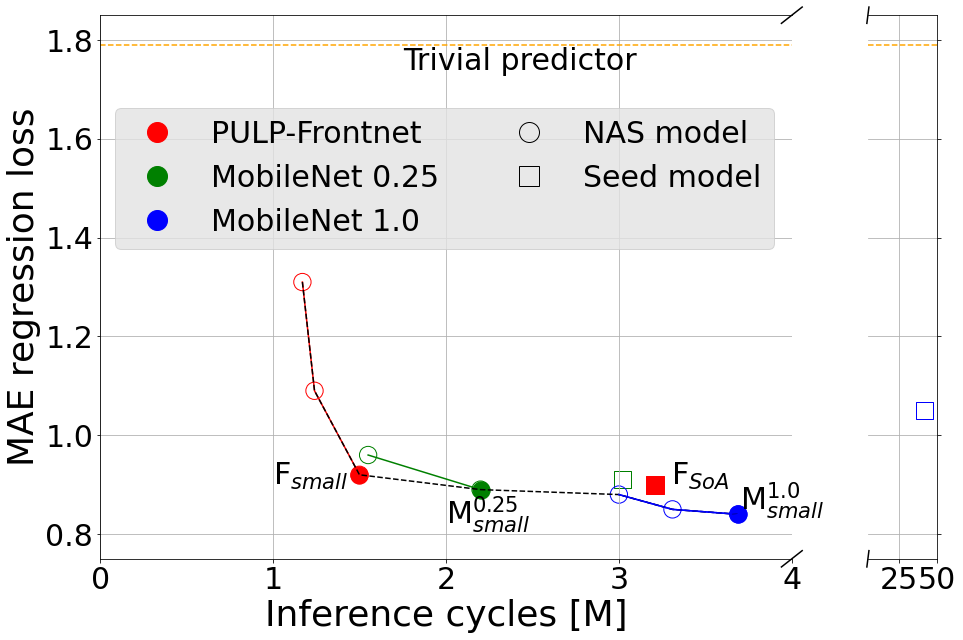}
  \caption{Pareto curves of the networks extracted from the NAS in the clock cycles vs. MAE space (lower is better).}
  \label{fig:pareto}
\end{figure}

The CNN deployment is based on the open-source PULP-NN library~\cite{garofalo2019pulp}, which encompasses optimized 8-bits fixed-point arithmetic kernels.
In particular, it exploits the eight general-purpose cores of the \textsc{Cluster} of GAP8 SoC and the ISA-specific optimizations to maximize kernels' performance. 
Given a theoretical maximum of \SI{32}{MAC/cycle} (4 MACs can be executed thanks to the int8 SIMD for each of the 8 cores), PULP-NN is able to reach a peak of \SI{15.6}{MAC/cycle} for squared-size images.
Note that the library peak is obtained by perfectly re-using data in the register file, reducing the number of necessary loads and stores per each MAC. 
If no data is stored in the register file for data re-use, two loads would be required for MAC inputs and one store to save the result in memory, reducing the theoretical peak to \SI{8}{MAC/cycle}.
On the other hand, the PULP-NN kernels consider data to be stored in the shared L1 \SI{64}{\kilo\byte} memory, relegating their applicability to small layers.

To cope with this constraint, we employ a second open-source tool, DORY~\cite{burrello2020dory}.
This tool automatically produces template-based C code, which wraps the PULP-NN kernels, managing the different levels of memories of GAP8 (i.e., L1, L2, and the external RAM) and orchestrating the tensors' movements.
In particular, DORY exploits a tiling approach to separate layers into small nodes whose tensors can fit the L1 memory of the system. 
Thanks to this, the kernels can be directly executed in these nodes.
DORY produces indeed the C routines, which i) execute the kernels of the small nodes with tensors stored in L1, and ii) double-buffer the movements of tensors from L2 to L1, to always have data ready for kernel execution. 
Notice that since the DMA is not blocking, the calls to the kernels are always overlapped with the asynchronous DMA calls.

\textbf{Closed-loop control.}
We deploy our models as part of the same closed-loop control system from~\cite{frontnet}, which allows to maintain the drone at a desired 3D pose in front of the person. 
Four main components are involved: \textit{i)} the CNN model outputs a point-in-time pose estimate from a single image, \textit{ii)} a Kalman filter produces smooth temporally-consistent sequences of poses, and \textit{iii)} a velocity controller computes the drone's desired pose from the subject's current pose and generates velocity setpoints to bring the drone to the desired pose.
\textit{iv)} the Crazyflie's low-level control, responsible for motor actuation and stabilization.
We adopt a Kalman filter decoupled between the model's four outputs, by assuming diagonal process and observation noise covariance matrices.
The velocity controller is also decoupled between linear velocity control, whose goal is to reach the specified target position, and angular velocity control, which keeps the subject centered in the image frame.
\section{Results} \label{sec:results}

\begin{table}[t]
    \begin{center}
      \caption{Test set experiment results}
      \label{tab:testset_results}
      \renewcommand{\arraystretch}{1.25}
      \begin{tabular}{lcccccccc} 
        \toprule
        \textbf{Network} & \multicolumn{4}{c}{\textbf{MAE}} & \multicolumn{4}{c}{\textbf{R2 score}} \\
        \cmidrule(lr){2-5}
        \cmidrule(lr){6-9}
         & $x$ & $y$ & $z$ & ${\phi}$ & $x$ & $y$ & $z$ & ${\phi}$\\
        \midrule
        \textbf{Trivial} & 0.46 & 0.61 & 0.17 & 0.55 & 0.0 & 0.0 & 0.0 & 0.0\\
        \textbf{F$_{\mathit{SoA}}$~\cite{frontnet}} & 0.19 & 0.18 & 0.09 & 0.44 & 0.80 & 0.65 & 0.55 & 0.26\\
        \textbf{F$_{\mathit{small}}$}  & 0.20 & 0.20 & 0.09 & 0.44 & 0.79 & 0.65 & 0.51 & 0.24\\

        \textbf{M$^{0.25}_{\mathit{small}}$}  & \textbf{0.17} & 0.20 & 0.08 & 0.44 & \textbf{0.84} & 0.63 & 0.62 & 0.22\\
        \textbf{M$^{1.0}_{\mathit{small}}$}  & 0.17 & \textbf{0.18} & \textbf{0.07} & \textbf{0.42} & 0.83 & \textbf{0.67} & \textbf{0.62} & \textbf{0.28}\\        
        \bottomrule
      \end{tabular}
    \end{center}
  \end{table}

\subsection{NAS Pareto analysis} \label{sec:pareto}

In Figure~\ref{fig:pareto}, we show and analyze the architectures found by our NAS algorithm. 
We compare on one axis the inference latency (number of clock cycles), whereas, on the other, we report the mean absolute error (MAE), expressed as the sum of L1 errors on ($x, y, z, \phi$) between the networks predicted values and the ground truth.
The \textit{trivial predictor}, i.e., a network that always predicts each output as its mean value in the test set, represents a lower bound to the models' MAE.
From our three seed network architectures, the NAS search discovers eight new models, most of which lie on the global Pareto front in the space of MAE vs. the number of cycles.
The found architectures range from 1.27M to 3.69M execution cycles, with corresponding MAE values from 1.31 to 0.84.
In detail, PULP-Frontnet models  occupy the left-most section of the Pareto curve, being very lightweight but less accurate.
The middle is populated by models derived from the MobileNet 0.25$\times$ seed.
Finally, the most accurate architectures are derived from MobileNet 1.0$\times$. This seed architecture is too big to fit in the GAP8's L2 memory.
However, our NAS algorithm can shrink it enough to find great solutions: absolute top-performing models that only increase latency by $15\%$ compared to $F_{SoA}$~\cite{frontnet}. 

From the global Pareto front, we select four models to deploy and further analyze.
The first is $F_{SoA}$, which corresponds to the original PULP-Frontnet, the current state-of-the-art, and our baseline.
The $F_{small}$ architecture is the fastest model, still achieving a MAE roughly equivalent to $F_{SoA}$.
$M_{small}^{1.0}$ is the most accurate architecture found by the NAS, but also the most expensive in terms of latency.
$M_{small}^{0.25}$ represents the most balanced trade-off between the two metrics, with both better than or equivalent to $F_{SoA}$.
In the following sections, we further analyze the selected architectures' performance on the test set and their behavior in the field in a closed-loop system.

\subsection{Regression performance} \label{sec:static_dataset}

\begin{figure}[t]
  \includegraphics[width=0.95\columnwidth]{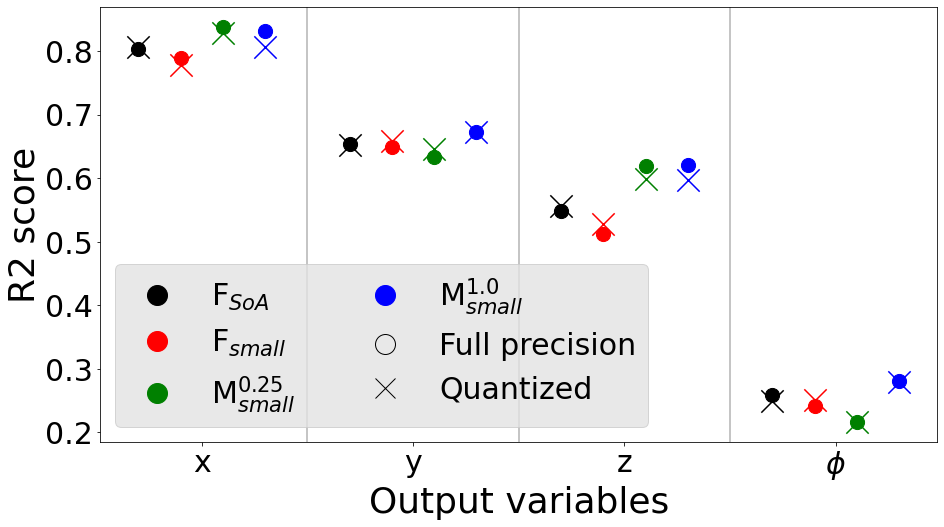}
  \caption{R2 score of the four deployed models (higher is better).}
  \label{fig:R2_score}
\end{figure}

In Table~\ref{tab:testset_results} and Figure~\ref{fig:R2_score}, we break down the models' regression performance in the four output variables, $(x, y, z, \phi)$.
For each output, in addition to the MAE values, we report the coefficient of determination $R^2$, a standard adimensional metric that represents the fraction of variance in the target variable explained by the model\footnote{$R^2 = 1 - \nicefrac{\sum_i(y_i - \hat{y}_i)^2}{\sum_i (y_i - \bar{y})^2}$ with $y_i$ the ground-truth output and $\hat{y}_i$ the model prediction for each test sample $i$, $\bar{y}$ the mean of ground-truth outputs. An $R2 = 1.0$ corresponds to a perfect regressor, while the trivial regressor achieves $R2 = 0.0$.\
}.
Compared to the MAE, the $R^2$ score quantifies the quality of a regressor independently of the target variable's variance and is, therefore, better suited for comparing regression performance between different variables.
Performance on all outputs shows trends consistent with those on the aggregated loss, confirming $M_{small}^{1.0}$ as the top performing on all outputs except for $x$, where $M_{small}^{0.25}$ performs slightly better.
In Figure~\ref{fig:R2_score}, we see that all models can predict $x$ and $y$ best while $\phi$ is the most difficult to estimate, matching the findings of background work.

\subsection{Onboard performance assessment} \label{sec:onboard_performance}

\begin{figure}[t]
  \includegraphics[width=\columnwidth]{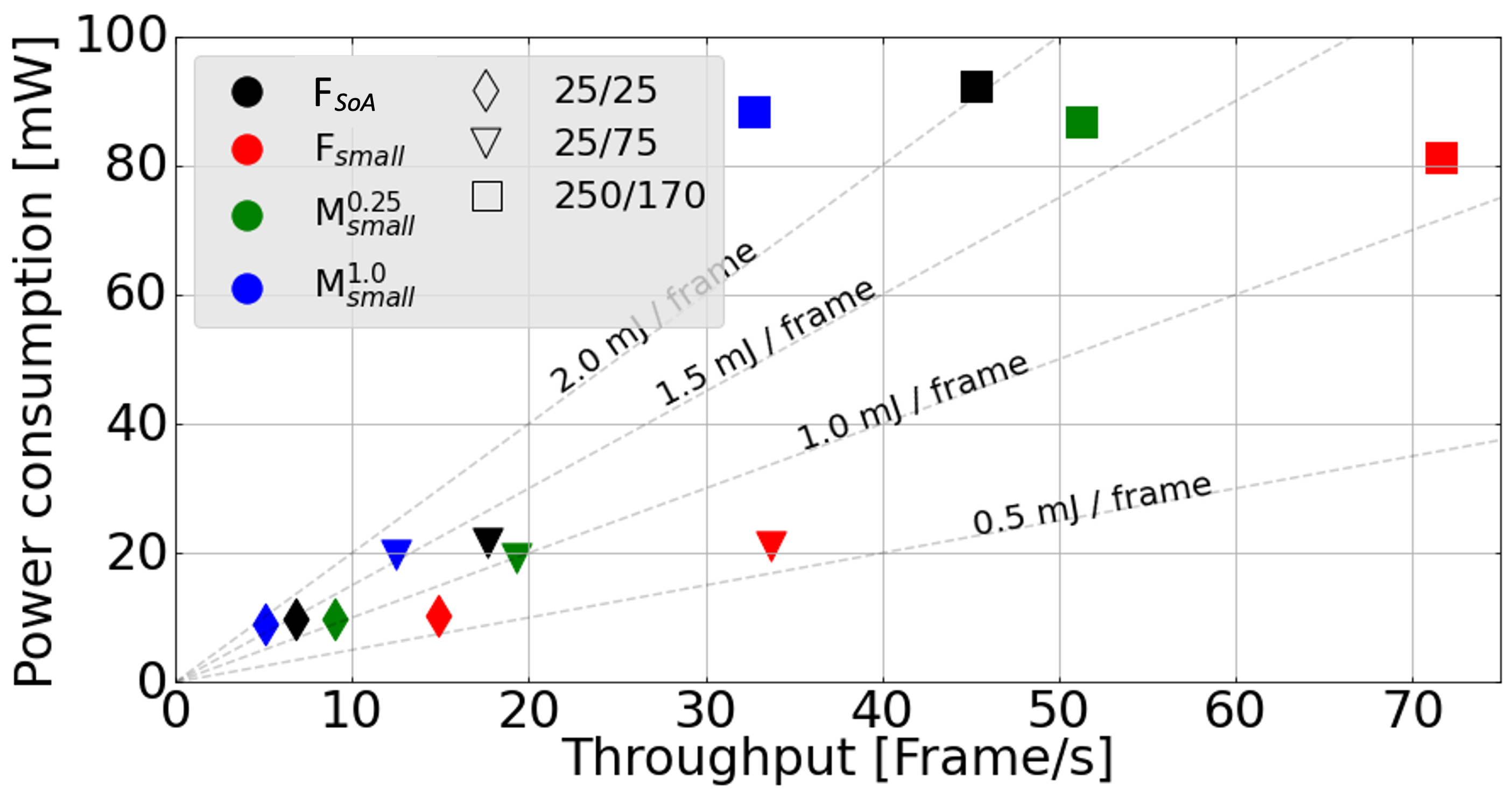}
  \caption{Throughput vs. power consumption for the four models at various operative points.}
  \label{fig:throughput_vs_power}
\end{figure}

\begin{figure}[t]
    \centering
    \includegraphics[width=0.9\columnwidth]{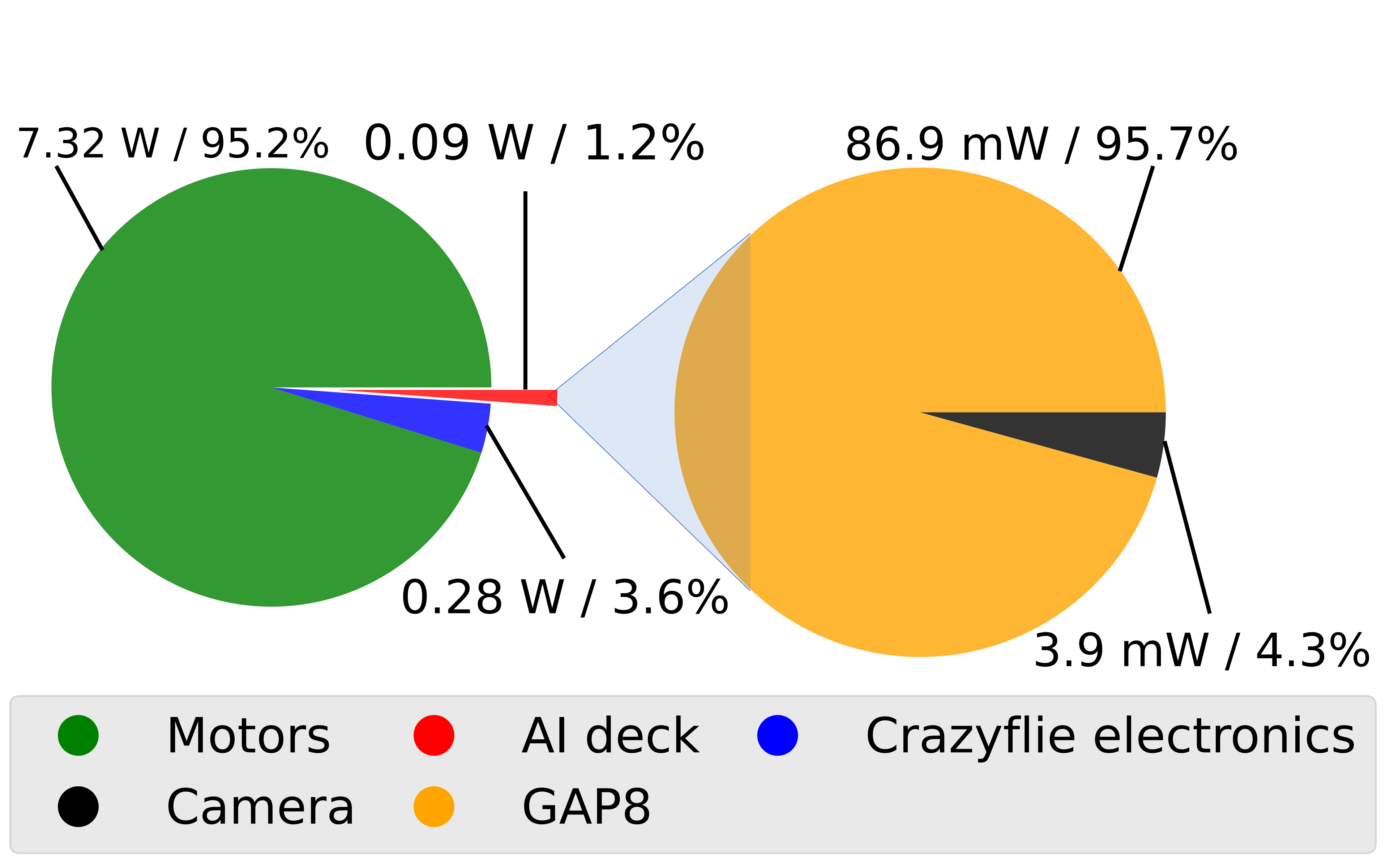}
    \caption{Nano-drone's power breakdown, running the M$_{small}^{0.25}$ model.}
    \label{fig:power-breakdown}
\end{figure}

We assess the onboard performance of the CNNs selected in the previous Section~\ref{sec:pareto}.
To nail down the computational, power, and memory requirements, we profiled our models, running them on the GAP8 SoC and using a RocketLogger data logger~\cite{sigrist2016rocketlogger} (\SI{64}{\kilo sp\second}). 
For these experiments, we test three SoC operating points: minimum power consumption with FC@\SI{25}{\mega\hertz} CL@\SI{25}{\mega\hertz} VDD@\SI{1}{\volt}, most energy-efficient with FC@\SI{25}{\mega\hertz} CL@\SI{75}{\mega\hertz} VDD@\SI{1}{\volt}, and maximum performance with FC@\SI{250}{\mega\hertz} CL@\SI{170}{\mega\hertz} VDD@\SI{1.2}{\volt}, as shown for the PULP-Frontnet baseline~\cite{frontnet}.
Table~\ref{tab:CNNs_profiling} summarizes the analysis on the four CNNs, showing that NAS models  significantly reduce  parameters (up to -85\% w.r.t. $F_{SoA}$), MACs, clock cycles, and memory (all three around -50\%). Power and throughput figures for the operative point used in the in-field experiments (max. performance) are also reported.
Figure~\ref{fig:throughput_vs_power} shows the relation between power and throughput, across the three operative points. We see that the two smaller models $M_{small}^{0.25}$ and $F_{small}$ improve upon the baseline $F_{SoA}$ in both regards, resulting in higher energy efficiency.
Figure~\ref{fig:power-breakdown} breaks down the power usage of the entire system. 
Crazyflie electronics plus the AI-deck cost 4.8\% of the total budget, almost saturating the power that can be allocated to sensing and computing~\cite {wood12pico}. 
As the GAP8 accounts for 24\% of that, there is a clear benefit in best optimizing its workload.

\begin{table}[t]
    \begin{center}
      \caption{Computation and memory footprint for inference on one frame (\textbf{F}: PULP-Frontnet, \textbf{M}: MobileNet).}
      \label{tab:CNNs_profiling}
      \renewcommand{\arraystretch}{1.3}
      \resizebox{\columnwidth}{!}{
      \begin{tabular}{lcccccc} 
        \toprule
        \textbf{Network}&\textbf{Params}&\textbf{MAC}&\textbf{Cycles}& \textbf{Memory} & \textbf{P [mW]} & \textbf{T [fps]}\\
        \midrule
        \textbf{F$_{\mathit{SoA}}$ \cite{frontnet}} & \SI{304}{\kilo\nothing} & \SI{14.7}{\mega\nothing} & \SI{3.2}{\mega\nothing} & \SI{499}{\kilo\byte} & 92.2 & 45.3\\
        \textbf{F$_{\mathit{small}}$}  & \SI{44}{\kilo\nothing} & \SI{7.6}{\mega\nothing} & \SI{1.5}{\mega\nothing} & \SI{231}{\kilo\byte} & 81.3 & 71.6\\
        \textbf{M$_{\mathit{small}}^{0.25}$}  & \SI{65}{\kilo\nothing} & \SI{7.4}{\mega\nothing} & \SI{2.2}{\mega\nothing} & \SI{591}{\kilo\byte} & 86.9 & 51.2\\
        \textbf{M$_{\mathit{small}}^{1.0}$} & \SI{54}{\kilo\nothing} & \SI{12.4}{\mega\nothing} & \SI{3.7}{\mega\nothing} & \SI{415}{\kilo\byte} & 88.3 & 32.7\\
        \bottomrule
      \end{tabular}
     }
    \end{center}
  \end{table}

\subsection{In-field experimental evaluation} \label{sec:infield}

We further validate the proposed networks in a closed-loop, in-field experiment.
We ask a human subject to follow a predefined path, while the drone is tasked to stay in front of them at eye level, at a distance of $\Delta = \SI{1.3}{\meter}$.
For consistency, we adopt the same test setup proposed by~\cite{frontnet}, with a \SI{50}{\second} path composed of 8 phases of increasing difficulty for the model (walking straight along different directions, walking along a curve, and rotating in place).
To ensure repeatability between different runs of the experiment, we ask the subject to completely ignore the drone's behavior and instead synchronize each step to the beats of a metronome.

The experiment is performed with both a subject and an environment outside of our training set, stressing the models' generalization capabilities.
We perform four experiment runs for each of the four models, plus one baseline run in which the drone is controlled using perfect information about the subject's pose from a mocap system, a total of 17 test flights.

Table~\ref{tab:infield_results} summarizes our results in this experiment.
We provide quantitative measures of three aspects of the system's performance: overall path completion, inference accuracy, and control accuracy.
We measure path completion for each model in terms of total flight time over the four runs and mean percentage of path completed, interrupting a run as soon as the person completely leaves the camera field of view.
In the challenging never-seen-before environment used for this experiment, the two PULP-Frontnet models struggle to complete the path (especially $F_{small}$, which reaches only 35\% on average), while the two MobileNet models consistently maintain the tracking until the end. We mark metrics corresponding to incomplete runs with an asterisk because they cannot be directly compared with other runs.

Inference accuracy then evaluates the models in isolation, measuring their ability to correctly estimate the subject's position w.r.t. to the drone. 
As in the reference work, here we do not consider the $z$ component because the target height is approximately constant in our task.
We report the MAE, to allow direct comparison with offline performance on the test set reported in Section~\ref{sec:static_dataset}. 
As expected, absolute performance decreases significantly due to the different environments.
At the same time, we see that the best in-field regression performance comes from $M_{small}^{0.25}$, instead of $M_{small}^{1.0}$ as on the test set. 
One explanation is the lower number of parameters in the former model, which makes it less prone to overfitting and thus generalize better.

\begin{figure}[t]
    \centering
    \includegraphics[width=\columnwidth]{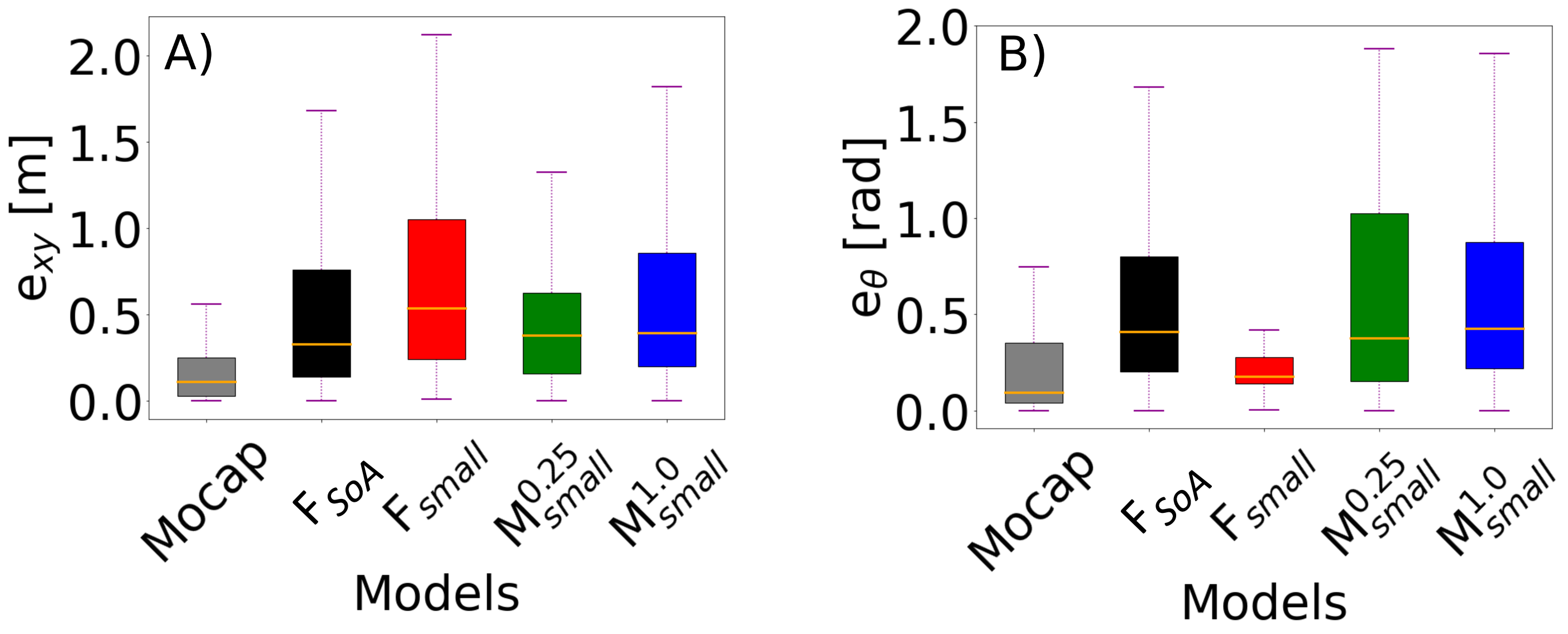}
    \caption{In-field control errors distribution (lower is better). Boxplot whiskers mark the $5^{th}$ and $95^{th}$ percentile of data.}
    \label{fig:infield_control}
\end{figure}

Finally, control accuracy evaluates the whole system's precision in tracking the subject as it moves along the path.
We compare the drone's actual pose against the desired pose, measuring two errors: the absolute position error $e_{xy}$ (i.e., the distance between the two poses in the horizontal plane) and the absolute angular error $e_\theta$ (i.e., the difference in orientation).
$M_{small}^{0.25}$ is the best on both metrics, confirming itself as the best-performing in-field.
In Figure~\ref{fig:infield_control}, $M_{small}^{0.25}$ is the model with the lowest variance in absolute position error, further explaining its better performance.
In addition, visually inspecting the drone's behavior shows that the $M_{small}^{0.25}$ model is noticeably more accurate than the baseline $F_{SoA}$.
To complement our results, we provide a supplementary video of the four model's behavior in the in-field experiments.


\begin{table}[t]
    \begin{center}
      \caption{In-field experiment results}
      \label{tab:infield_results}
      \resizebox{\columnwidth}{!}{
      \renewcommand{\arraystretch}{1.25}
      \begin{tabular}{lccccccc} 
        \toprule
        \multirow{2}[3]{*}{\textbf{Network}} & \multirow{2}[3]{*}{\shortstack[c]{\textbf{Flight}\\\textbf{time [s]}}} & \multirow{2}[3]{*}{\shortstack[c]{\textbf{Completed}\\ \textbf{path [\%]}}} & \multicolumn{3}{c}{\textbf{MAE}} & \multicolumn{2}{c}{\textbf{Mean pose error}} \\
        \cmidrule(lr){4-6} \cmidrule(lr){7-8}
        & & & $x$ & $y$ & ${\phi}$  & $e_{xy}$ [m] & $e_\theta$ [rad]\\
        \midrule
        \textbf{Mocap} & 165 & 100 & 0.0 & 0.0 & 0.0 & 0.18 & 0.21\\
        \textbf{F$_{\mathit{SoA}}$~\cite{frontnet}} & 140 & 85 & 0.33* & 0.12* & 0.77* & 0.72* & 0.78*\\
        \textbf{F$_{\mathit{small}}$}  & 58 & 35 & 0.81* & 0.53* & 0.55* & 0.65* & 0.42*\\
        \textbf{M$^{0.25}_{\mathit{small}}$}  & \textbf{165} & \textbf{100} & \textbf{0.25} & \textbf{0.11} & \textbf{0.52} & \textbf{0.49} & \textbf{0.59}\\
        \textbf{M$^{1.0}_{\mathit{small}}$} & \textbf{165} & \textbf{100} & 0.31 & 0.13 & 0.52 & 0.58 & \textbf{0.59}\\
        \bottomrule
      \end{tabular}
      }
    \end{center}
  \end{table}

\section{Conclusion} \label{sec:conclusion}

This system paper presents a practical use case on NAS technologies applied for a challenging robotic visual perception human pose estimation task on nano-drones.
Starting from two seed CNNs (i.e., PULP-Frontnet and MobileNetv1), we select four Pareto-optimal models to be deployed aboard a resource-constrained (i.e., sub-\SI{100}{\milli\watt}) nano-quadrotor.
We assess the capabilities of the CNNs with a thorough analysis: from their regression performance on a disjoint test set, an onboard performance evaluation (power consumption and throughput), down to an in-field closed-loop test in a \textit{never-seen-before} environment.
Our best model improves the SoA by reducing the in-field control error of 32\% with a real-time inference rate of $\sim$\SI{50}{\hertz}@\SI{90}{\milli\watt}.

\bibliographystyle{./IEEEtran}
\bibliography{biblio}

\end{document}